\documentclass[journal]{IEEEtran}

\usepackage{ifpdf}

\ifCLASSINFOpdf
   \usepackage[pdftex]{graphicx}

\else

\fi

\usepackage{amsmath}

\usepackage{algorithmic}

\usepackage{multirow}
\usepackage{algorithm}
\usepackage{booktabs}

\usepackage{array}

\usepackage{url}
\usepackage{capt-of}
\usepackage{wrapfig}
\usepackage{rotating}
\usepackage[acronym]{glossaries}
\usepackage{multirow}

\begin{document}

\newcolumntype{C}[1]{>{\centering\arraybackslash}p{#1}}

\title{GANsfer Learning: Combining labelled and unlabelled data for GAN based data augmentation}
%
%
%

\author{{Christopher Bowles, Roger Gunn, Alexander Hammers, Daniel Rueckert}
\thanks{C. Bowles and D. Rueckert are with the Biomedical Image Analysis Group, Department of Computing, Imperial College London, UK.}
\thanks{A. Hammers is with the Guy's and St Thomas' PET Centre, King’s College London, UK}
\thanks{R.Gunn is with Imanova Ltd., London, UK and the Department of Medicine, Imperial College London, UK.}}

%



\maketitle
\newacronym{mri}{MRI}{Magnetic Resonance Imaging}
\newacronym{mr}{MR}{Magnetic Resonance}
\newacronym{pet}{PET}{Positron Emission Tomography}
\newacronym{gan}{GAN}{Generative Adversarial Network}
\newacronym{ml}{ML}{Machine Learning}
\newacronym{spect}{SPECT}{Single Photon Emission Computed Tomography}
\newacronym{ct}{CT}{Computed Tomography}
\newacronym{pnfa}{PNFA}{Progressive Non-Fluent Aphasia}
\newacronym{sd}{SD}{Semantic Dementia}
\newacronym{lpa}{LPA}{Logopenic Aphasia}
\newacronym{mnd}{MND}{Motor-neuron Disease}
\newacronym{pdd}{PDD}{Parkinson's Disease with Dementia}
\newacronym{wmhpvo}{WMH$_{\mathrm{pvo}}$}{White Matter Hyperintensity of Presumed Vascular Origin}
\newacronym{rssi}{RSSI}{Recent Small Subcortical Infarct}
\newacronym{adni}{ADNI}{Alzheimer’s Disease Neuroimaging Initiative}
\newacronym{adni_go}{ADNI-GO}{Alzheimer’s Disease Neuroimaging Initiative - Grand Opportunity}
\newacronym{adni_2}{ADNI-2}{Alzheimer’s Disease Neuroimaging Initiative - 2}
\newacronym{adni_3}{ADNI-3}{Alzheimer’s Disease Neuroimaging Initiative - 3}
\newacronym{adni_1}{ADNI-1}{Alzheimer’s Disease Neuroimaging Initiative - 1}
\newacronym{fdg}{FDG}{fluorodeoxyglucose}
\newacronym{dsi}{DSI}{Disease State Index}
\newacronym{oasis}{OASIS}{Open Access Series of Image Studies}
\newacronym{ab42}{A$\beta _{42}$}{amyloid-$\beta$ 1-42}
\newacronym{ad}{AD}{Alzheimer's Disease}
\newacronym{flair}{FLAIR}{Fluid-attenuated Inversion Recovery}
\newacronym{svd}{SVD}{Small Vessel Disease}
\newacronym{wmh}{WMH}{White Matter Hyperintensity}
\newacronym{wm}{WM}{White Matter}
\newacronym{gm}{GM}{Grey Matter}
\newacronym{rf}{RF}{Radio-Frequency}
\newacronym{csf}{CSF}{Cerebrospinal Fluid}
\newacronym{dwi}{DWI}{Diffusion Weighted Imaging}
\newacronym{fmri}{fMRI}{Functional Magnetic Resonance Imaging}
\newacronym{mra}{MRA}{Magnetic Resonance Angiography}
\newacronym{3d}{3D}{3-dimensional}
\newacronym{2d}{2D}{2-dimensional}
\newacronym{4d}{4D}{4-dimensional}
\newacronym{ftd}{FTD}{Frontotemporal Dementia}
\newacronym{bvftd}{bvFTD}{Behavioural Variant Frontotemporal Dementia}
\newacronym{dlb}{DLB}{Dementia with Lewy bodies}
\newacronym{vd}{VD}{Vascular Dementia}
\newacronym{fad}{FAD}{Familial Alzheimer's Disease}
\newacronym{dti}{DTI}{Diffusion Tensor Imaging}
\newacronym{auc}{AUC}{Area Under the Curve}
\newacronym{roi}{ROI}{Region of Interest}
\newacronym{ms}{MS}{Multiple Sclerosis}
\newacronym{pca}{PCA}{Posterior Cortical Atrophy}
\newacronym{msssim}{MS-SSIM}{Multi-scale structural similarity}
\newacronym{em}{EM}{Expectation Maximisation}
\newacronym{cnn}{CNN}{Convolutional Neural Network}
\newacronym{vae}{VAE}{Variational Autoencoder}
\newacronym{swd}{SWD}{Sliced Wasserstein Distance}
\newacronym{uqi}{UQI}{Universal Quality Index}
\newacronym{lncc}{LNCC}{Local Normalised Cross-correlation}
\newacronym{ncc}{NCC}{Normalised Cross-correlation}
\newacronym{mi}{MI}{Mutual Information}
\newacronym{j}{J}{Jaccard Coefficient}
\newacronym{assd}{ASSD}{Average Symmetric Surface Distance}
\newacronym{hd}{HD}{Hausdorff Distance}
\newacronym{icc}{ICC}{Intra Class Correlation}
\newacronym{sse}{SSE}{Sum of Squared Errors} 
\newacronym{ssd}{SSD}{Sum of Squared Differences}
\newacronym{rpc}{RPC}{Reproducibility Coefficient}
\newacronym{cv}{CV}{Coefficient of Variation}
\newacronym{lstlga}{LST-LGA}{Lesion Growth Algorithm}
\newacronym{lstlpa}{LST-LPA}{Lesion Prediction Algorithm}
\newacronym{lst}{LST}{Lesion Segmentation Toolbox}
\newacronym{malpem}{MALPEM}{Multi-Atlas-Label Propagation with Expectation-Maximisation based refinement}
\newacronym{ia}{IA}{Image Analogies}
\newacronym{pggan}{PGGAN}{Progressive Growing of GANs}
\newacronym{wgan}{WGAN}{Wasserstein Generative Adversarial Network}
\newacronym{dcgan}{DCGAN}{Deep Convolutional Generative Adversarial Network}
\newacronym{mse}{MSE}{Mean Squared Error}
\newacronym{snr}{SNR}{Signal to Noise Ratio}
\newacronym{dsc}{DSC}{Dice Similarity Coefficient}
\newacronym{js}{JS}{Jaccard Similarity}
\newacronym{crf}{CRF}{Conditional Random Field}
\newacronym{svm}{SVM}{Support Vector Machine}
\newacronym{gmm}{GMM}{Gaussian Mixture Model}
\newacronym{mni}{MNI}{Montreal Neurological Institute}
\newacronym{ffd}{FFD}{Free Form Deformation}
\newacronym{apoe}{APOE}{Apolipoprotein E}
\newacronym{mci}{MCI}{Mild Cognitive Impairment}
\newacronym{lmci}{LMCI}{Late Mild Cognitive Impairment}
\newacronym{emci}{EMCI}{Early Mild Cognitive Impairment}
\newacronym{cn}{CN}{Normal Controls}
\newacronym{tsne}{tSNE}{t-Distributed Stochastic Neighbour Embedding}
\newacronym{cdr}{CDR}{Clinical Dementia Rating}

\begin{abstract}
Medical imaging is a domain which suffers from a paucity of manually annotated data for the training of learning algorithms. Manually delineating pathological regions at a pixel level is a time consuming process, especially in \gls{3d} images, and often requires the time of a trained expert. As a result, supervised machine learning solutions must make do with small amounts of labelled data, despite there often being additional unlabelled data available. Whilst of less value than labelled images, these unlabelled images can contain potentially useful information. In this paper we propose combining both labelled and unlabelled data within a \gls{gan} framework, before using the resulting network to produce images for use when training a segmentation network. We explore the task of deep grey matter multi-class segmentation in an \gls{ad} dataset and show that the proposed method leads to a significant improvement in segmentation results, particularly in cases where the amount of labelled data is restricted. We show that this improvement is largely driven by a greater ability to segment the structures known to be the most affected by \gls{ad}, thereby demonstrating the benefits of exposing the system to more examples of pathological anatomical variation. We also show how a shift in domain of the training data from young and healthy towards older and more pathological examples leads to better segmentations of the latter cases, and that this leads to a significant improvement in the ability for the computed segmentations to stratify cases of \gls{ad}.
\end{abstract}


%
\IEEEpeerreviewmaketitle

\section{Introduction}

\IEEEPARstart{D}{ata} augmentation is the process of expanding a training dataset by including additional data derived from the available real data. In the case of image data, such approaches can include the rotation, translation, reflection and the spatial and intensity scaling of the images, as well as the application of complex deformations. The aim is to increase the number of training samples by introducing additional feasible data points, and thereby  
reduce the potential for overfitting a model trained on the data. For example, in a small dataset it is likely that some irrelevant information such as the angle of an image is coincidentally correlated with an image label. Learning this correlation will cause the learning algorithm to erroneously classify images based on angle. By artificially rotating the training images and including these copies during training, this coincidental correlation is removed and the learning algorithm is forced to discover alternative discriminative features. 

It has been noted~\cite{deform} that data augmentation is not extensively used in neuroimaging applications, primarily due to the extensive preprocessing options available to remove a much of the variance which augmentation methods aim to introduce. Registration to a standard space obviates rotation, spatial scaling and translation augmentation, while intensity normalisation makes intensity scaling augmentation unnecessary. Meanwhile, random deformations are difficult to realistically model in a space where anatomical constraints such as symmetry and rigidity must be respected.

Recent work~\cite{Frid-Adar2018,SPIE,salehinejad2017generalization,Bowles2018} has shown that \glspl{gan}~\cite{Goodfellow2014} can be used to produce labelled medical images of high enough quality to perform learning on, and that these can be used to effectively conduct data augmentation. \Glspl{gan} are a class of neural networks which aim to learn to produce images with the characteristics of those contained in a given training distribution. A \gls{gan} consists of two components. The generator maps a random vector to an image, while the discriminator takes an image and outputs a belief as to whether it's real or synthetic. During training, the discriminator is provided a batch of real and synthetic images to learn from, with the loss function being the cross entropy loss. The generator then produces a new batch of synthetic images and is updated according to the discriminator's loss on this batch. This process is repeated until convergence. Many developments to this standard framework have since been proposed, as well as offshoots including conditional \glspl{gan} which introduced discriminator losses to tasks such as style transfer~\cite{zhu2017unpaired} and super-resolution~\cite{Ledig2016}.

The original \gls{gan} was built on with \glspl{dcgan}~\cite{Radford2015} where training stability and image size and quality were improved by modifying the two network architectures to remove fully connected layers and replace pooling layers with (fractional) strided convolutions. Some drawbacks of the early \gls{gan} formulations were the lack of a true image quality related loss function, and there being a need to balance the amount of training cycles provided to the two networks. These problems were addressed in~\cite{Arjovsky2017}, where the Jenson-Shanon divergence approximating formulation of the previous work was replaced with one in which the discriminator approximates the Wasserstein distance instead. More recently, \gls{pggan}~\cite{nvidia} was proposed. This multi-resolution approach to training was shown to reliably generate images up to 1024-by-1024px, well beyond the maximum of 256-by-256px seen in previous formulations. This is achieved by progressively growing the size of the networks, starting with a small \gls{gan} generating 4-by-4px images, and adding layers to both the generator and discriminator throughout training, successively doubling the output image size until the desired size is reached. The choice of optimal \gls{gan} architecture for a given task is non-trivial. Work in~\cite{lucic2017gans} showed that, averaged across datasets under optimal hyperparameter selection, the performance of many \gls{gan} formulations are not statistically different from one another. The choice of ``best \gls{gan} formulation" may therefore be dependant on how easily these optimal hyperparameters can be found for a given dataset. 

While \glspl{gan} have been shown have uses across the medical imaging domain, their use for data augmentation is still relatively unexplored. In~\cite{Frid-Adar2018}, synthetic liver lesions are generated using a \gls{dcgan} architecture and used to augment a dataset for the purpose of lesion classification. A similar approach is taken in~\cite{SPIE} where synthetic normal and abnormal chest radiographs are generated using two \gls{dcgan}-like architectures, and  in~\cite{salehinejad2017generalization}, synthetic chest X-rays are generated in order to balance an imbalanced dataset. 
All of these papers report that using \glspl{gan} for data augmentation leads to improvements in classification accuracy for their respective tasks. This approach was also applied to segmentation tasks~\cite{Bowles2018}, yielding an improvement in segmentation accuracy when using small datasets. 

In this paper, we consider whether this approach can be improved by introducing additional unlabelled data. One of the limitations of training \glspl{gan} on small amounts of data is that the manifold learned by the generator will be characterised by a small number of modes, with only images with subtle variations coming from around these modes being produced. While these slight variations have so far proved sufficient to improve performance in classification and segmentation tasks, there is scope to improve upon this. To be able to learn a smooth manifold
, a critical mass of data is required. However, if such a large amount of labelled data is available, there is likely little need to perform augmentation and using a \gls{gan} becomes redundant. Instead, we wish to move towards learning this smooth manifold from a substantially smaller number of labelled images. We propose to do this by leveraging a large amount of unlabelled images in addition to the limited labelled images, using a technique inspired by transfer learning.

Transfer learning has proved to be an effective approach to neural network training across many applications, including in medical imaging~\cite{hoo2016deep}. It involves training a model on a separate, usually very large, dataset, then applying it to the desired problem for which there is comparatively little available data. This is usually done with a form of fine tuning, where all or part of the pre-trained network is subsequently trained on samples from the target problem. The intuition behind this is that many low level features will be shared across tasks. At the lowest levels these might be looking for simple edges and colours, while slightly higher level features might be activated by line segments. It is only deeper in the network that these features take on any meaning relating to the target domain and, even deeper, specific task. It can therefore be beneficial to learn the low level features from a separate dataset. 
This has several advantages including reducing computation time due to model reuse, and increasing the quality of the learned features. 
This process can be thought of as learning (or reusing) a general purpose feature extractor, followed by learning how to interpret these features in the context of the task at hand. Feature extraction and interpretation are decoupled and learned separately. We propose a similar framework for training a \gls{gan} where we decouple the learning of structural variation and the learning of the appearance of these structures. In the case of labelled images, appearance encompasses not only pixel intensities in image space, but also the corresponding binary segmentation labels contained in additional channels. The aim is to train the parts of the network which correspond to appearance from a small amount of labelled images, while the parts responsible for generating structural variation are trained on a large unlabelled dataset.

Recent work~\cite{madani2018semi} showed that incorporating unlabelled data using a \gls{gan} framework can lead to significant improvements in accuracy in a chest X-ray classification task. 
Another approach was presented in~\cite{ross2018exploiting} for endoscopic video segmentation where a conditional \gls{gan} using a UNet~\cite{UNet}-like generator is first trained on the unlabelled data to perform an arbitrary auxiliary task - in this case an image re-colourisation procedure. This pre-trains the first half of the UNet as an effective feature extractor by leveraging the unlabelled data. The labelled data is then used to fine-tune the UNet in the target segmentation task. An increase in classification accuracy from 51\% to 73-76\% when labelled data is most limited in~\cite{madani2018semi}, and in \gls{dsc} (a measure of overlap between 0 (no overlap) and 1 (complete overlap)) from 0.57 to 0.65 in~\cite{ross2018exploiting} shows the potential for incorporating unlabelled data into a learning system.

\section{Methods}

\subsection{Rationale}

\begin{figure}[h!]
  \centering
  \includegraphics[width=1\linewidth]{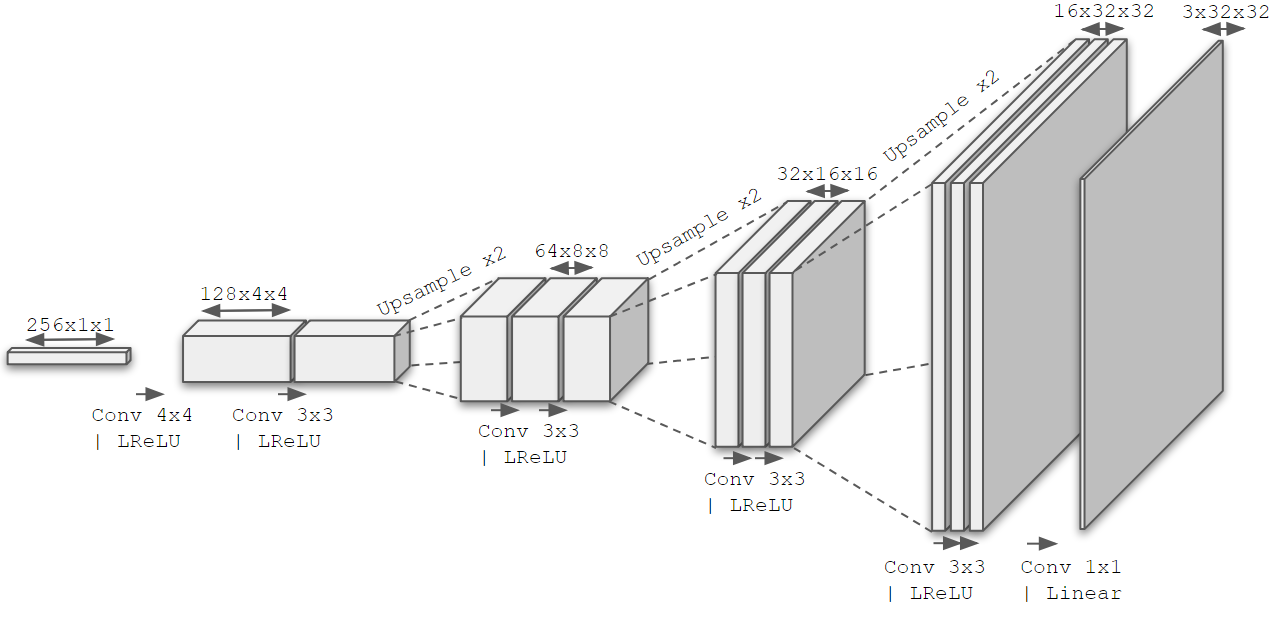}
\caption{Architecture of a typical \gls{pggan} generator for a 3 channel 32-by-32px image from a 256 element latent vector.}
\label{fig:PGGAN_arch}
\end{figure}

We first consider the structure of a \gls{gan}, particularly the generator, to understand which layers are responsible for defining the anatomical structures, and which are used to generate their appearance. For the purposes of these experiments we use the \gls{pggan}~\cite{nvidia} algorithm as it was found reliably generate multi-channel images at required image size.  
A possible final generator architecture for a small network is shown in Figure~\ref{fig:PGGAN_arch}. By considering the final convolution process we see that each of the output channels is a weighted sum of the feature maps from the previous layer. This layer is therefore responsible for producing both the image and segmentation channels from the feature maps provided by the penultimate layer. 

This suggests that segmentation maps are only generated in the final layers from a feature space shared with the image channel. This also follows when we consider the \gls{pggan} grows over time during training, with the earlier layers tasked with generating low resolution images, and therefore the low frequency information such as the anatomy, and the later layers responsible for the higher frequency information such as texture. 
Having established that the final layers are responsible for the appearance, and therefore the segmentation maps, and that the earlier layers are responsible for the anatomical variance, we can consider how to train these components separately. A traditional transfer learning approach would dictate training on the large unlabelled dataset first, before refining the final layers on the small labelled dataset. However, this would involve increasing the number of output channels at the refinement stage, thereby not only changing the desired output distribution but also its dimensionality. This sudden change in the discriminator objective function was found to result in a failure to train and subsequent network collapse. We hypothesise that this is due to the required information to form the segmentations not being available in the final layers
Instead, we propose to pre-train the \gls{gan} using the small labelled dataset, and refine the early layers using the large unlabelled dataset, before fine-tuning using a combination of both. This avoids radically changing desired output distribution, leading to a smooth transition between phases. 
The details of this process are given below.

\subsection{GANsfer Learning}

\emph{Phase 1: Train the \gls{gan} until using labelled data.} 

The goal of this phase is to pre-train the generator ($\mathbf{G}$) and condition it to ensure the necessary information to generate segmentations is present in the final layers. After this step, image diversity will be low due to limited training data, however image and segmentation quality will be high.

\emph{Phase 2: Fix the weights of the final layers of $\mathbf{G}$ and train a new discriminator using the unlabelled data and the image channel from $\mathbf{G}$.}

The goal of this phase is to learn more anatomical variation from the unlabelled data. Having learned a set of weights in the final layers which produce segmentation maps, we increase image diversity by fixing these layers and continuing to train on the unlabelled data. As the real data are now only single channel images, and the trained discriminator expects a multi-channel image, we replace this with a new discriminator ($\mathbf{D_I}$). $\mathbf{D_I}$ takes as input the real training images and the synthetic image channel from the output of $\mathbf{G}$. To allow $\mathbf{D_I}$ to catch up with $\mathbf{G}$, regular \gls{gan} training is performed using a ratio of 100 $\mathbf{D_I}$ updates to 1 $\mathbf{G}$ update for the first 5 training cycles, mirroring the pattern in~\cite{Arjovsky2017}.

\emph{Phase 3: Add another discriminator $\mathbf{D_S}$ trained using an equal mix of the segmentation channels of labelled images and of the generated images after \textit{phase 2} as ground truth, and the segmentation channels from $\mathbf{G}$. Perform regular \gls{gan} training by updating $\mathbf{G}$ according to the loss from both $\mathbf{D_S}$ and $\mathbf{D_I}$. Gradually unfreeze layers of $\mathbf{G}$ up to the final layer.}

The goal of this phase is to improve the quality of the generated images. The previous phase taught $\mathbf{G}$ to produce greater anatomical variation, but at a cost of image quality due to the frozen layers. In this phase the frozen layers are gradually unfrozen, allowing these weights to be re-optimised with respect to the earlier layers. This is similar to the original \gls{pggan} training procedure. This phase can therefore be thought of as a second pass through $\mathbf{G}$, optimising each layer in turn to smoothly reattach the newly trained early layers to the final layers. The final layer remains frozen to ensure the image and segmentation channels remain coupled. During this phase the \gls{gan} is trained using two discriminators. $\mathbf{D_I}$ is retained from \textit{phase 2} and ensures $\mathbf{G}$ retains its ability to produce varied images. The second, $\mathbf{D_S}$, is a newly trained network which is trained purely on the segmentation channels and ensures the quality of the generated segmentation channels is preserved. Its training set consists of the segmentation channels from the labelled dataset, combined in equal parts with the segmentation channels from a new dataset generated using $\mathbf{G}$ at the end of \textit{phase 2}. This ``self-teaching" ensures $\mathbf{G}$ does not to forget variation it learned previously. Without this, there would be a significant disparity in the anatomical variation contained within the image dataset and segmentation dataset. Under these circumstances $\mathbf{D_I}$ and $\mathbf{D_S}$ would be in conflict, with $\mathbf{D_I}$ encouraging greater variation, while $\mathbf{D_S}$ encourages reduced variation.

\begin{figure}[h!]
  \centering
  \includegraphics[width=0.6\linewidth]{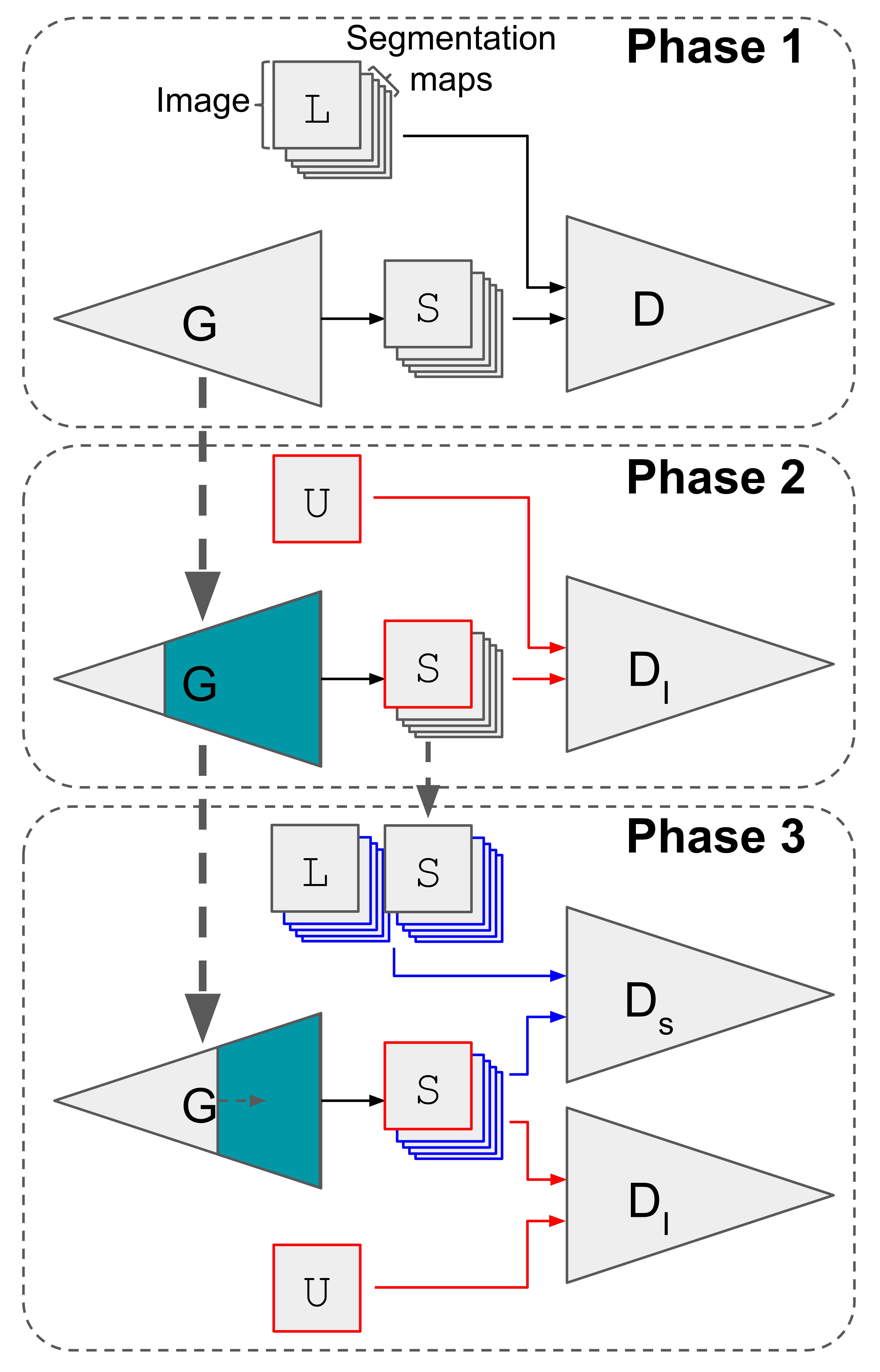}
\caption{The three phases of GANsfer learning. \textit{Phase 1} trains the whole network with generator (G) and discriminator (D) on labelled data (L) to produce synthetic data (S). \textit{Phase 2} trains the early layers of the generator using unlabelled data (U) and a new image based discriminator (D$_\mathrm{I}$). \textit{Phase 3} reintroduces the later layers using a combination of both labelled data, unlabelled data and previously synthesised images. It uses combined feedback from the image based discriminator and a new segmentation based discriminator (D$_\mathrm{S}$).}
\label{fig:GANsfer}
\end{figure}

In summary, \textit{phase 1} ensures the final layers of the generator contain feature maps which can be linearly combined to produce both image and segmentation channels. \textit{Phase 2} trains the early layers to generate increased anatomical variability. \textit{Phase 3} refines the generator to improve image quality. The data produced by the generator can then be used for data augmentation. This whole process is shown in Figure~\ref{fig:GANsfer}.

\section{Experiments}

Experiments were devised to evaluate what effect the proposed method has when different amounts of labelled data are available. We investigate the task of multi-class deep grey matter segmentation on $T_1$-weighted \gls{mr} images with 7 anatomical structures. This application was chosen for three main reasons. Firstly, it is a typical medical image segmentation task, and as such, results on this task should provide an indication of performance on similar tasks. Secondly, data for multi-class problems are particularly time consuming to manually annotate, and therefore provide a realistic use case for GANsfer learning. Finally, the relatively small region of interest means that the GANs required can be trained to the full resolution of the images in a reasonable amount of time, allowing for an extensive investigation with 5 fold cross-validation.

\subsection{Data}

For the labelled dataset we use data provided for the MICCAI 2013 Grand Challenge on Multi-Atlas Labelling\footnote{data available at: \url{www.synapse.org/#!Synapse:syn3193805/wiki/217780}}. This contains 35 images from the OASIS-1 dataset, manually annotated by Neuromorphometrics, Inc\footnote{\url{www.neuromorphometrics.com}}. Each image is accompanied by clinical information including age, gender and \gls{cdr} - a 5 point rating scale for \gls{ad} consisting of: healthy (\gls{cdr} 0), very mild \gls{ad} (0.5), mild \gls{ad} (1), moderate \gls{ad} (2) and severe \gls{ad} (3). Of the 35 images, 5 are repeated from the same subject, and are discarded. The remaining 30 images (Age: 18-90, median 25; Gender: 20F/10M; 29 healthy, 1 very mild \gls{ad}) are affinely co-registered to a standard space with a 1mm isotropic voxel grid and intensity normalised to a zero-mean unit-variance across all non-background voxels, after which an 80-by-80-by-60px region of interest, defined in common space and covering the deep grey matter structures, is extracted. These images are divided into 5 folds for cross-validation, each containing 24 training and 6 testing images, with each image contributing 60 \gls{2d} 80-by-80px axial slices.

Each slice has corresponding label information indicating the segmentations of the: accumbens, amygdala, caudate, hippocampus, pallidum, putamen and thalamus, in the form of 7 binary image channels. We found that preprocessing these segmentation channels to make them more closely correlated with \gls{mr} channel led to improved image quality and increased coupling between the \gls{mr} and segmentation channels. To preprocess each segmentation channel, we transfer the pixel intensities from the \gls{mr} channel into the corresponding region in the segmentation channel, and subtract an estimated \gls{wm} intensity. These values are then inverted if necessary to remain positive and used as the segmentation channels for \gls{gan} training. 


A key purpose of this is to remove the sharp edges present in the binary segmentation channels. To generate these would require some of feature maps in the penultimate generator layer to have sharp edges, which could not also be used to generate the \gls{mr} channel. This would lead to a potential decoupling of the parts of the network responsible for generating the \gls{mr} and segmentation channels. To preserve the ability for the network to produce realistic segmentations after further training with unlabelled data, it is important that these processes are tightly linked. Towards this end, it is also important that the contrast levels in the segmentation channels are similar to the corresponding regions in the \gls{mr} channel. 
All structures share a border with the \gls{wm}, so it is therefore appropriate to use \gls{wm} intensity as a base from which to calculate the contrast of each structure.

The unlabelled dataset consists of the entire OASIS-1 dataset. This contains 436 images (Age: 18-96, median 54; Gender: 268F/168M images; 336 healthy, 70 very mild \gls{ad}, 28 mild \gls{ad}, 2 moderate \gls{ad}), of which 20 are repeated scans of the same healthy subjects. These images are preprocessed in the same way as described above. This dataset has a much older age profile than the labelled dataset, and contains many more examples of \gls{ad} pathology. We hypothesise that by incorporating this older, more pathological, data into the \gls{gan}, the resulting network will produce more examples with features associated with old age and \gls{ad}. This will then lead to more accurate segmentations of subjects with these features.

\subsection{Postprocessing}

\begin{figure}[h]
  \centering
  \includegraphics[trim={13cm 7cm 13cm 1cm},clip,width=\linewidth]{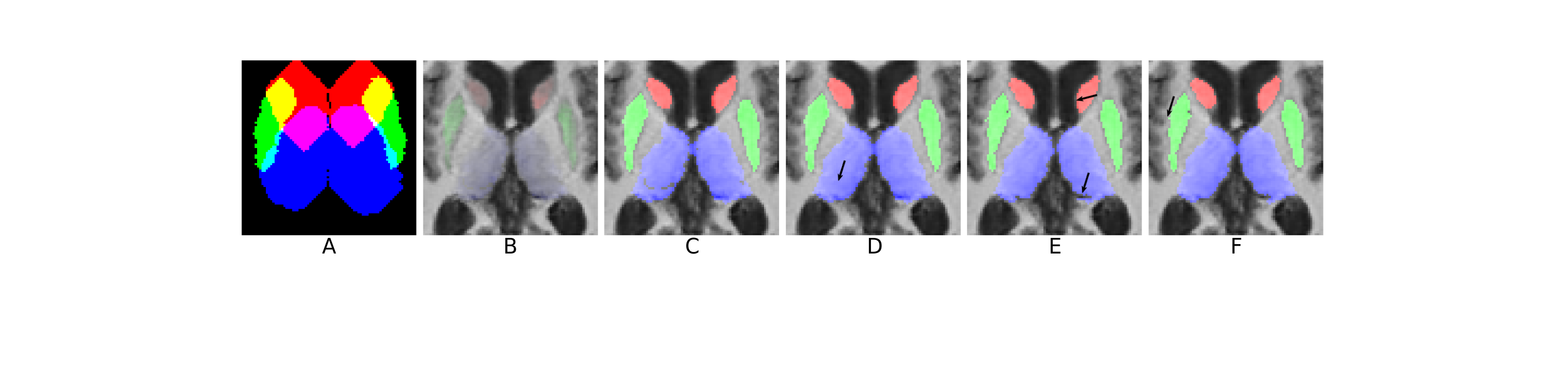}
\caption{Postprocessing visualised on 3 segmentation channels. A) Mask derived from real segmentations. B) Masked image. C) Binarised segmentation channels. D) Holes filled. E) Intensity based threshold applied. F) Holes filled and spurs removed. Arrows indicate the effect of each step.}
\label{fig:PostProcAnot}
\end{figure}

Since the \gls{gan} is trained to produce non-binary segmentation channels, a set of postprocessing steps are required to binarise them and correct minor errors. First, each image is assigned a slice number through nearest neighbour comparison of the \gls{mr} channel to the real training images. A \gls{3d} mask is then defined for each segmentation channel containing all points within 10mm of the union of all available labelled training images for that experiment. The appropriate slice from this \gls{3d} mask is then used to remove any obviously incorrect segmentations from each generated image. Each segmentation channel is then binarised  and any resulting holes removed through morphological closing and hole filling. The \gls{mr} intensity distribution within each structure is then calculated, with any pixels falling outside of the mean $+/-$ 2 standard deviations removed from the segmentation. Finally, any new holes are filled and morphological opening removes any small disconnected components and spurs. This whole process can be seen in Figure~\ref{fig:PostProcAnot}.

\begin{figure}[h!]
  \centering
  \includegraphics[trim={3cm 3cm 3cm 5cm},clip,width=0.8\linewidth]{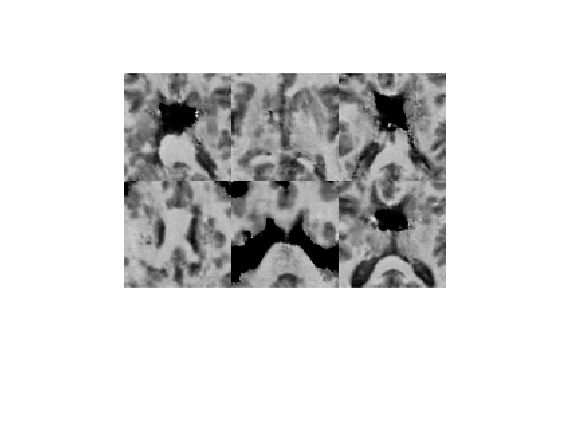}
\caption{Sample of 3 unrealistic generated images removed from the dataset.}
\label{fig:Purged}
\end{figure}

As well as post-possessing, the generated images are also filtered based on image quality. Whilst generated images were generally found to be of reasonable quality, the generator can occasionally produce unrealistic images. To filter these out, a score for each image defined as the minimum Euclidean distance between the \gls{mr} channel of each generated image and any image from the full dataset is found. These are ranked and the generated images with scores above the 75$^{th}$ percentile are removed. This is a very conservative threshold to ensure no unrealistic images are kept, while removing realistic images is not detrimental as more can be generated at very little cost. Examples of the highest scoring images are shown in Figure~\ref{fig:Purged}.

\subsection{Segmentation network}

To assess the quality of the synthetic data, we propose to train a segmentation network with and without synthetic labelled data added to the available real labelled data. We use DeepMedic~\cite{CRF} for this purpose. DeepMedic is a general purpose segmentation network which has been shown to give good results in a variety of segmentation tasks. We modify the default architecture from \gls{3d} to \gls{2d} as described in~\cite{CRF}, and otherwise use default hyper-parameters and settings, including left/right reflection augmentation. Segmentations are evaluated using \gls{dsc}. The primary metric used for comparison is an overall \gls{dsc}, treating all 7 tissue classes as foreground, though we also examine the results for individual structures.

\subsection{Impact of quantity of available labelled data}

For every learning task there is a threshold beyond which additional labelled data provides negligible improvement. The goal of augmentation is therefore to lower the amount of data required to reach this optimal performance. Baseline experiments using 1, 3, 6, 12 and 24 labelled training images show no significant improvement between 12 and 24 images (see Figure~\ref{fig:Ratios1} later for full results), indicating the that this optimal level of performance has been reached. The aim is therefore to achieve results closer to this level using fewer labelled images and augmenting with synthetic data. Experiments were therefore performed by making 1, 3 and 6 labelled images available, taken from the pool of 24 allocated training images at each fold. Despite not expecting any significant performance increase on the baseline results when 12 and 24 labelled images are used, a subset of experiments are also performed at these levels to better understand the effects of synthetic data.

\subsection{Impact of quantity of synthetic data}

The quality of synthetic data produced by the \gls{gan} will never be as high as real manually labelled images. We must therefore consider the relative exposure the segmentation network should get to the real and synthetic pools of training data. This is done by allowing the segmentation network to sample from each pool of data with a different probability, effectively allowing for different ratios between real and synthetic data to be explored. We consider 4 ratios: 100:1, 10:1, 2:1 and 1:1. For example, a ratio of 100:1 means that for every 100 patches sampled from real data, 1 patch is sampled from synthetic data. Note that the pool of synthetic images available is effectively infinite, hence it is highly unlikely to sample the same image twice, while the pool of real images is relatively small, meaning the same region is likely to be sampled multiple times during training. 

\subsection{Pre-training with large amounts of data}
\begin{figure}[h!]
  \centering
  \includegraphics[width=1\linewidth]{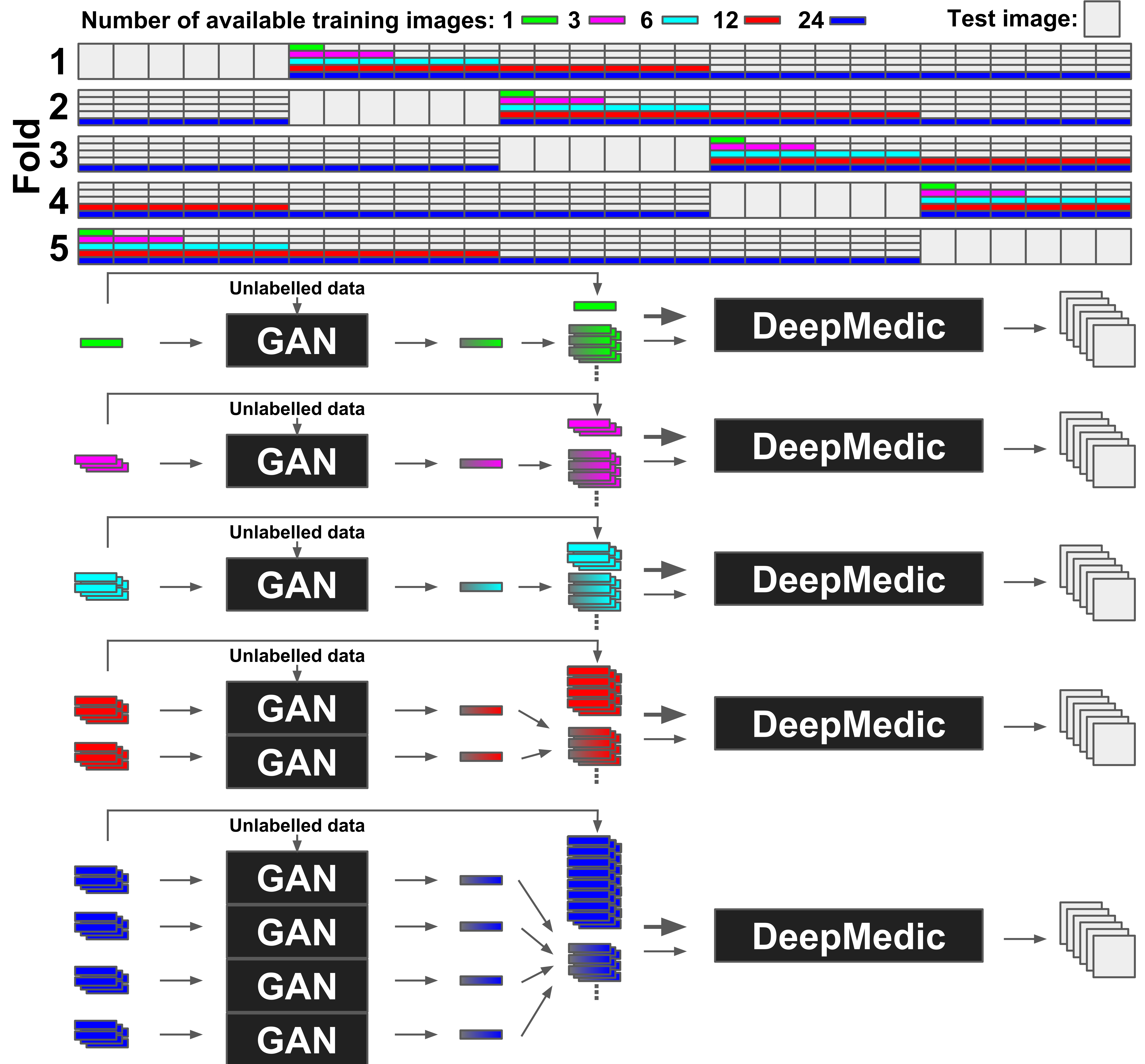}
\caption{An overview of the experimental setup. At the top, the 30 labelled images are divided into training and test sets for 5 folds. For each fold, the training set is further divided to simulate cases where 1, 3, 6, 12 or all 24 images are available for training. Underneath, the process of training the required GANs and DeepMedic networks to investigate each level of available labelled data (1, 3, 6, 12 and 24, colour coded as above) for a single fold is shown. The available labelled and unlabelled data is first used to train a \gls{gan} using GANsfer learning. The generator is then used to create a synthetic dataset. A DeepMedic network is then trained by sampling (with varying probabilities) from the real and synthetic data, with the resulting model used to segment the 6 test images for that fold. Note that in the case of 12 (red) and 24 (blue) labelled images, multiple \glspl{gan} are trained on blocks of 6 images, rather than training a single \gls{gan} on the all the images.}
\label{fig:experiments}
\end{figure}

Early experiments indicated that pre-training on fewer (1, 3 or 6) labelled images produces higher quality images than when pre-training on more (12 or 24) images. Though more variation was observed when more images were used, the appearance would suffer. This can be attributed to the \gls{gan} attempting to perform interpolation between exemplar images - behaviour which is not exhibited when few images are used, with the \gls{gan} generating images from around these modes with only small variations. Once a sufficient amount of training images is provided, the \gls{gan} begins to attempt to produce images with greater variation. This is normal and usually desired \gls{gan} behaviour, however it is counter-productive in \textit{phase 1}, where we desire high quality images and variation is unimportant. This perceived loss of quality is confirmed when we perform segmentation experiments using only synthetic images after \textit{phase 1}. Experiments on a single fold show an average overall DSC of 0.68, 0.73, 0.73, 0.68 and 0.64 when training on 1, 3, 6, 12 and 24 images respectively. To avoid this behaviour having a negative impact on the final results in the cases where 12 or 24 labelled images are available, the training data in these cases is further split into groups of 6 images, and 2 or 4 \glspl{gan} respectively are trained using each group. After training, synthetic data from each \gls{gan} is combined and used to augment the real training data. Figure~\ref{fig:experiments} shows an overview of the experimental setup across the 5 folds.

\subsection{Full dataset analysis}

A consequence of having limited labelled training data is a similarly limited amount of test data upon which to directly test the trained segmentation models. While the 30 labelled images are sufficient to perform simple \gls{dsc} based comparisons between different quantities of available data, they do not facilitate a deeper analysis. By only having one mild \gls{ad} subject and 6 subjects with an age greater than 50, they do not provide a robust means of examining the effects on elderly and more pathological cases. We therefore perform further indirect evaluation on the unlabelled data by applying the trained segmentation models across the full dataset and using the segmentation volumes as features to build a classifier to differentiate between cases of very mild \gls{ad} and mild or moderate \gls{ad} (\gls{cdr} 0.5 and \gls{cdr} 1 or 2). 

After removing all repeated scans and training images, 287 healthy, 69 very mild \gls{ad}, 28 mild \gls{ad} and 2 moderate \gls{ad} subjects are available for analysis. Each image is segmented using five trained models (one per level of available data) and the structure volumes are extracted. These volumes form a 7-dimensional feature vector (one component for each tissue type) which is used to train a simple logistic regression classifier. Each experiment uses 5-fold cross-validation and is repeated 100 times. The accuracy and \gls{auc} metrics are calculated. 

For these experiments we use five models previously trained from a single fold. This fold was chosen as its training set does not contain the single AD subject and has the lowest average age of all folds (25). This lets us explore the scenario where only labelled images for young and healthy subjects are available, yet we wish to segment images of older and pathological subjects. We also perform the same analysis using segmentations provided by \gls{malpem}~\cite{MALPEM} to allow for further comparisons. \gls{malpem} is a \gls{3d} multi-atlas method to perform tissue segmentation, which uses the same set of 30 unique labelled images as used in our experiments, and has been applied successfully in \gls{ad} progression studies~\cite{ledig2018structural}.

\gls{malpem} offers significantly greater \gls{dsc} results than observed when using \gls{2d} DeepMedic. This difference is likely a result of \gls{malpem} being a \gls{3d} method, and therefore having access to more contextual information. Baseline results using the the two methods on a single fold show DeepMedic achieving an overall \gls{dsc} of 0.80, 0.85, 0.87 when using 1, 6 and 24 labelled images respectively, with \gls{malpem} achieving results of 0.80, 0.92 and 0.92 using the same subject atlases. The segmentations provided by \gls{malpem} using all training atlases can therefore be used as a surrogate for manual segmentations. The agreement between these and the computed segmentations can be analysed under the assumption that a greater overlap with \gls{malpem} segmentations would indicate greater accuracy. While not providing an absolute measure of performance, it does allow for further insights to be obtained using the full dataset. Using the same segmentations as used in the classification experiment above, we compute the overall and per-class \gls{dsc} using \gls{malpem} segmentations as reference. We then examine how subject age and \gls{cdr} classification affect the improvement seen when using synthetic data augmentation.

\section{Results and discussion}

\subsection{Ablation and optimisation}

To examine the impact of each stage of the proposed method we performed an ablation study where we measure results at the end of each phase of GANsfer learning. Using a single fold and one available labelled image, we performed segmentation using the generator output at the end of each stage. We evaluated the synthetic data on its own and combined in a ratio of 2:1 real to synthetic data. The results of this study can be seen in Table~\ref{tab:abl}.

\begin{figure}[h]
\small
\centering
\captionof{table}{\textbf{Ablation study:} \gls{dsc} observed on a single fold using one labelled training image at different stages during the \gls{gan} training pipeline. Results are given using synthetic images produced by the \gls{gan} at the end of each of the three training phases (\textit{P1, P2 and P3}), with (+) and without real data. The total \gls{dsc}, \gls{dsc} for each structure, and mean \gls{dsc} across all structures are provided. Baseline (BL) results using no synthetic data are also shown for reference.\label{tab:abl}}
\begin{tabular}{@{}p{0.8cm}lllllllll@{}}
\toprule
            & Total & Ac. & Am. & Ca. & Hi. & Pa. & Pu. & Th. & Avg \\ \midrule
\textit{BL}    & 80.1    & 46      & 57     & 80    & 58        & 78     & 81    & 84     & 69 \\
\textit{P1}     & 81.6    & 57      & 56     & 82    & 67        & 78     & 82    & 87     & 73 \\
\textit{P1}+    & 81.4    & 51      & 59     & 80    & 67        & 80     & 82    & 85     & 72 \\
\textit{P2}     & 79.7    & 24      & 51     & 79    & 67        & 72     & 75    & 84     & 64 \\
\textit{P2}+    & 82.0    & 49      & 60     & 81    & 68        & 81     & 80    & 87     & 72 \\
\textit{P3}     & 79.5    & 44      & 55     & 80    & 71        & 69     & 75    & 85     & 68 \\
\textit{P3}+    & 83.1    & 54      & 63     & 83    & 72        & 80     & 81    & 87     & 74 \\
\textit{P2}\&\textit{3}  & 79.0    & 38      & 53     & 80    & 68        & 71     & 73    & 85     & 67 \\
\textit{P2}\&\textit{3}+ & 83.9    & 56      & 62     & 84    & 74        & 81     & 81    & 88     & 75 \\ \bottomrule
\end{tabular}
\end{figure}

\begin{figure}[h]
  \centering
  \includegraphics[trim={0cm 0cm 0cm 0cm},clip,width=0.8\linewidth]{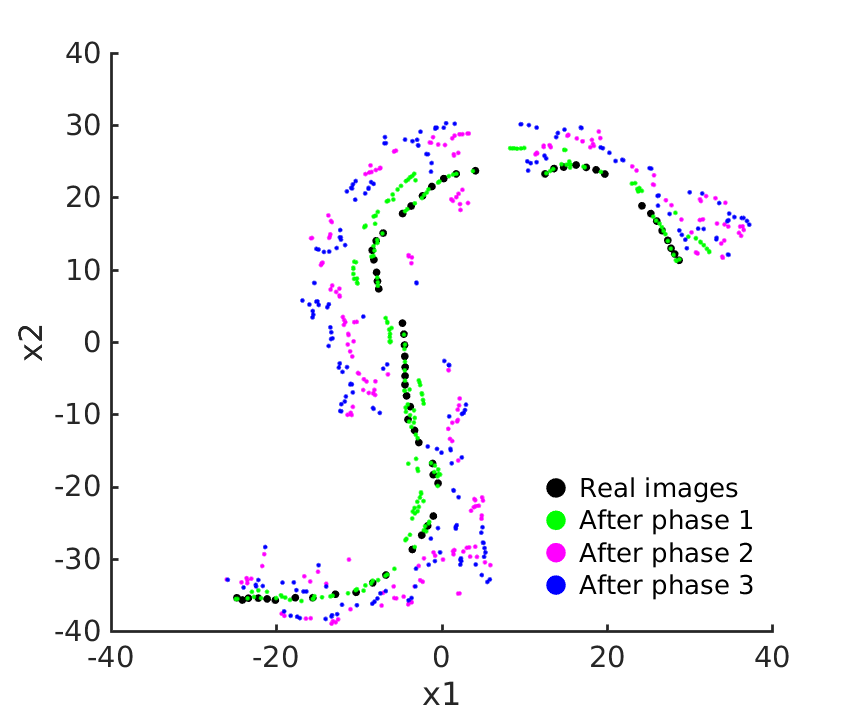}
\caption{\Gls{tsne} visualisation of training images, and the output from the same random selection of latent vectors after each training phase. The single training volume contributes 60 images and the output after \textit{phase 1} follows these closely. Images produced after \textit{phase 2} and \textit{phase 3} are further away, indicating greater variability. Axes $x1$ and $x2$ correspond to the embedding coordinates found through \gls{tsne}.}
\label{fig:TSNE}
\end{figure}

\begin{figure}[h!]
  \centering
  \includegraphics[trim={0cm 0cm 0cm 0cm},clip,width=1\linewidth]{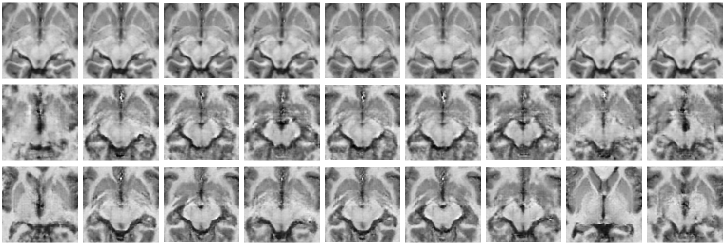}
  
  \vspace{1em}
  
  \includegraphics[trim={0cm 0cm 0cm 0cm},clip,width=1\linewidth]{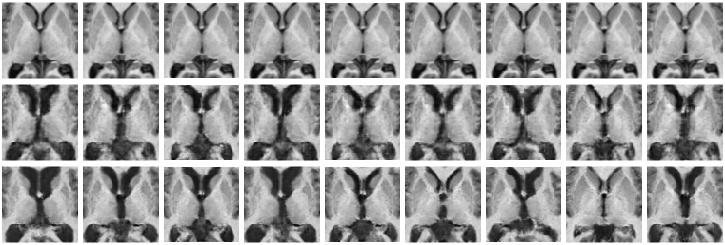}
\caption{\gls{gan} output after each phase of training covering two example regions. For each region, 9 latent vectors were found which map to approximately the same image after \textit{phase 1} (first row). The output from the same latent vectors were then generated after phases \textit{2} and \textit{3}. There is significantly more variation found after \textit{phase 2}, at the cost of lower quality images (second row). An improvement in quality, while maintaining variability, can then be seen in the output after \textit{phase 3} (third row).}
\label{fig:Abl1}
\end{figure}

\textit{Phase 1} is similar to the method used in~\cite{Bowles2018}, and does not involve any additional unlabelled data. It is therefore expected that we observe an improvement in segmentation performance. There is little difference when real data is used in addition to the synthetic data. This suggests that the generated images encompass all the relevant information from the real images, and are of sufficient quality to train from.

After \textit{phase 2}, using the synthetic data alone leads to worse results than observed after \textit{phase 1}, but when combined with real data, it produces better results. This can be attributed to the synthetic data now containing additional information having been exposed to the unlabelled dataset. There is, however, also a reduction in image quality, leading to a worse performance when used on its own. 
This reduction in quality is addressed in \textit{phase 3}, as evidenced by improved results compared to those after \textit{phase 2}. The best results were then found by combining the synthetic data produced after \textit{phase 2} and \textit{phase 3}. This improvement is mostly driven by better hippocampal segmentation. Since the hippocampus is known to be affected by \gls{ad}, it is possible that by using both sets of synthetic data, we include more examples of \gls{ad} pathology. All future experiments therefore use a combination of synthetic data from \textit{phases 2 and 3}.

The effects of each phase of training can also been visualised. Figure~\ref{fig:TSNE} shows a \gls{tsne}~\cite{maaten2008visualizing} visualisation of training images, and synthetic images after each phase of training. A \gls{tsne} embedding allows us to visualise high dimensional data in a low dimensional space. Points which are close in the high dimensional space will also be close in the low dimensional embedding and vice-versa. This gives an indication of variation, but not quality.

Figure~\ref{fig:Abl1} shows sample output for two regions, at the end of each phase of training. Whilst there is a reduction in image quality evident after \textit{phase 2}, this is rectified in \textit{phase 3} with no loss of variability apparent in either Figure~\ref{fig:TSNE} or \ref{fig:Abl1}.

\begin{figure}[h!]
  \centering
  \includegraphics[trim={0.1cm 17cm 1.1cm 0cm},clip,width=0.8\linewidth]{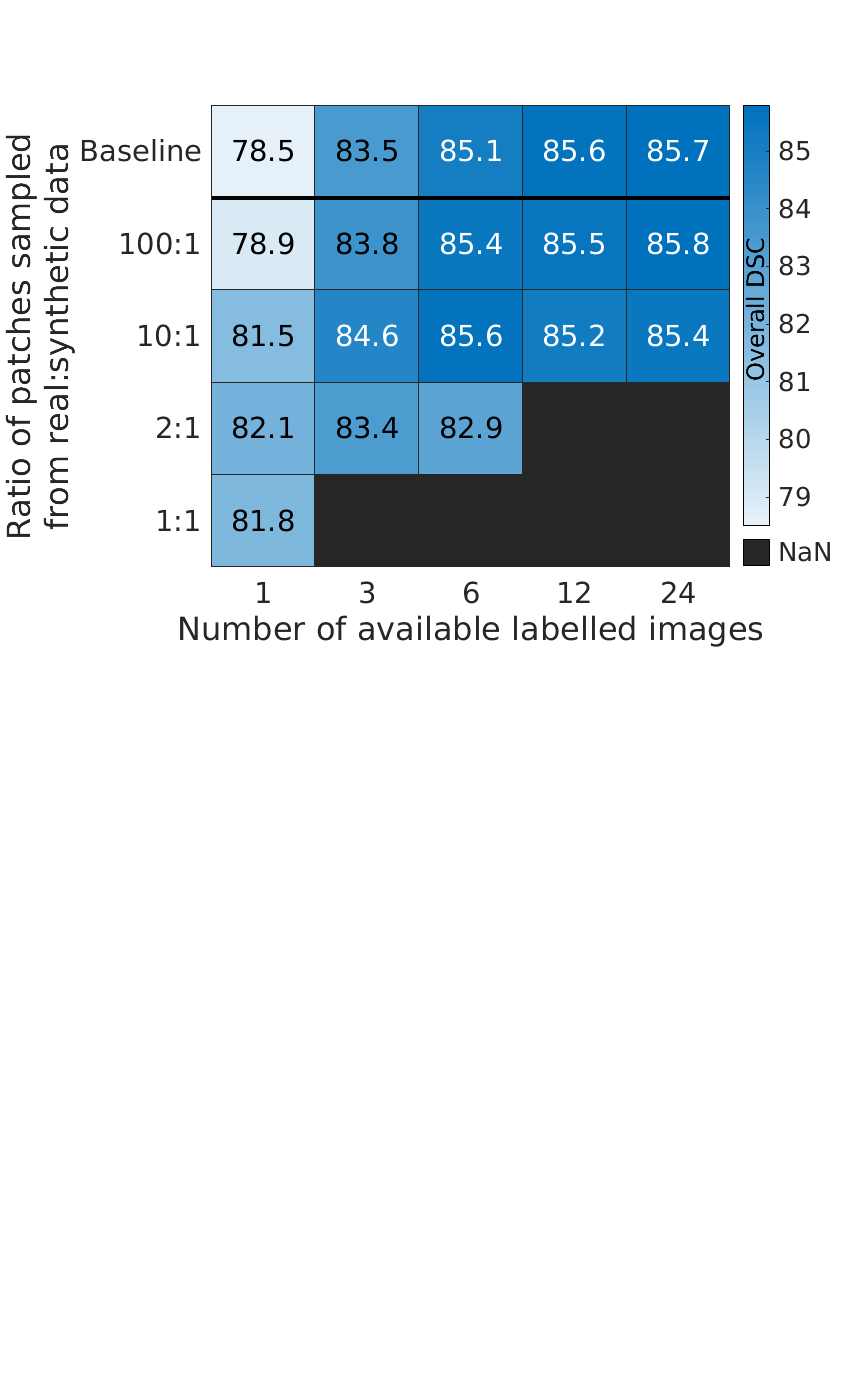}
\caption{Observed \gls{dsc} at baseline, and with different sampling rates of synthetic data during segmentation network training. There is a clear trend of more synthetic data being useful as the amount of real data is reduced.}
\label{fig:Ratios1}
\end{figure}

Results when sampling real and synthetic data at different rates during DeepMedic training can be seen in Figure~\ref{fig:Ratios1}. The results clearly show that more synthetic data is beneficial when less real data is available. This is expected, as when more real images are available, more variation is already present in the training set, and so the additional synthetic data will have less impact, to the extent that when 12 or more real images are available, there is no evidence of improvement at any ratio.

\subsection{Labelled dataset analysis}

\begin{figure*}[h]
  \centering
  \includegraphics[trim={7cm 0cm 7cm 0cm},clip,width=1\linewidth]{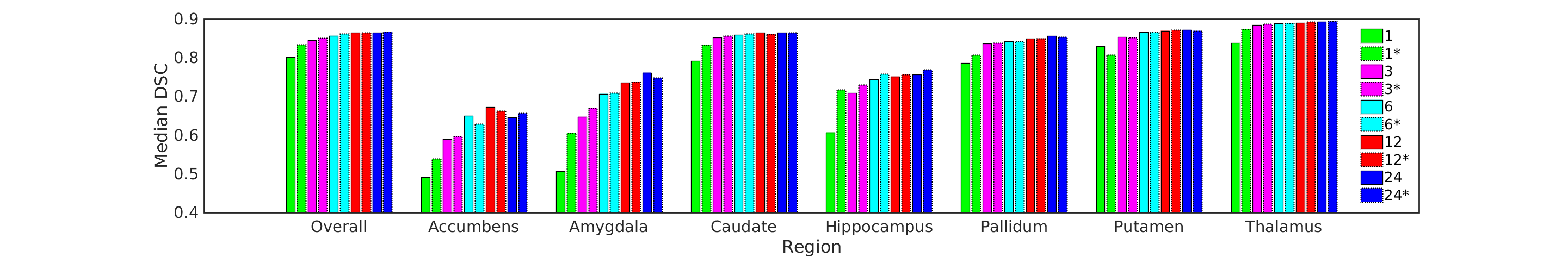}
\caption{Impact of using additional synthetic data on segmentation accuracy for each of the seven structures. Each pair of coloured bars shows the difference between the baseline results (left of pair) and results with the optimal amount of additional synthetic data (*, right of pair).}
\label{fig:RegionsNew}
\end{figure*}

\begin{figure}[h]
  \centering
  \includegraphics[width=0.8\linewidth]{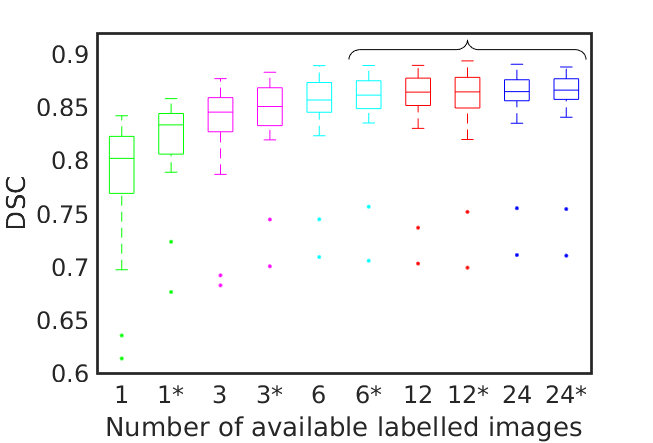}
\caption{Box plots showing the distribution of \gls{dsc} across all 30 images. Each coloured pair shows results with (*) and without synthetic data. Results within the bracket are not significantly different from each other at a 5\% significance level; all other result pairs are. The two outliers common to each experiment correspond to the eldest subject (lowest \gls{dsc}) and single very mild \gls{ad} subject (second-lowest \gls{dsc}).}
\label{fig:OverallBoxNew}
\end{figure}

Using the optimal ratios found previously, we now examine the results across the labelled dataset. Figure~\ref{fig:OverallBoxNew} shows the distribution of observed \gls{dsc} values with and without synthetic data at each level of available real data. Whilst we see significant improvements when 1 or 3 labelled images are available, neither of these are sufficiently large enough to score higher than using 3 or 6 images respectively. However, when 6 labelled images are available, we do see results which are not statistically different from those seen with 12 or 24 labelled images. It is reasonable to assume from the lack of improvement between baseline \glspl{dsc} for 12 and 24 images that this is approaching the maximum \gls{dsc} which can be achieved on this dataset using this segmentation algorithm. Observed \gls{dsc} on each of the seven deep grey matter structures with and without synthetic data can also be seen in Figure~\ref{fig:RegionsNew}. It is interesting to note that the largest improvements in \gls{dsc} can be seen in the hippocampus and amygdala, two structures which are known to be affected by \gls{ad}. This suggests that exposing the system to more examples of anatomical variation in \gls{ad} leads to an improvement in segmentation of these structures, even among a largely healthy cohort.

\subsection{Full dataset analysis}

\begin{figure}[h]
  \centering
  \includegraphics[trim={2cm 0cm 2cm 0cm},clip,width=0.8\linewidth]{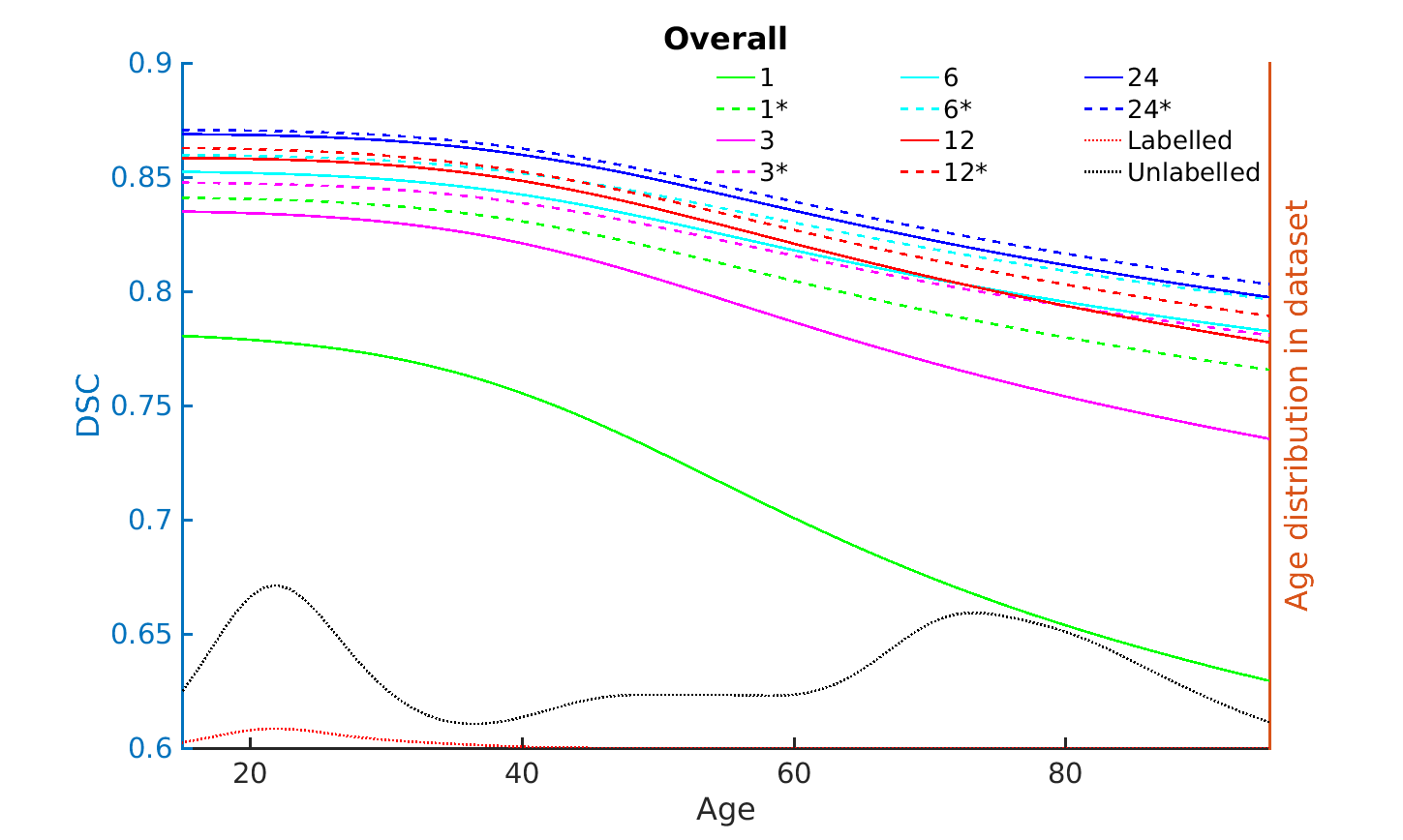}
\caption{Overall \gls{dsc} using \gls{malpem} segmentations for reference at different ages. Results for segmentations computed with (*) and without synthetic data augmentation for each level of available labelled images (1,3,6,12,24) are shown. The relative age distributions for the full labelled and unlabelled datasets are also shown. All data is smoothed using kernel regression to highlight the overall trends.}
\label{fig:AgeDSC}
\end{figure}

One benefit of using unlabelled data is that it extends the domain of training images from generally young and healthy to older subjects with more cases of \gls{ad}. The impact of this can therefore only be fully measured on the full dataset. This was done by using \gls{malpem} segmentations as a surrogate for ground truth labels. Figure~\ref{fig:AgeDSC} shows how the observed \gls{dsc} varies with subject age, with and without synthetic data, along with a visualisation of the distribution of ages within the labelled and unlabelled dataset. It is clear that in all cases, expected performance decreases as age increases, however, this effect is reduced when synthetic data is used. We can also see that in the cases where 12 and 24 labelled images are available there is little benefit to using synthetic data at the younger ages, agreeing with the results from the labelled data. However there does appear to be a larger improvement at the older end of the age spectrum. 

\begin{figure}[h]
\small
\centering
\captionof{table}{\label{tab:preds} Accuracy and \gls{auc} metrics comparing the ability of segmentation volumes to differentiate between \gls{cdr} 0.5 and \gls{cdr} 1 or 2 subjects when computed with (Aug.) and without (BL) augmentation, and when using \gls{malpem} (M), when different numbers of labelled images are available (\textit{N}). Results which are statistically different between corresponding baseline and augmentation results (2-tailed t-test, 5\% significance level) are shown in bold. Results which are not significantly different from ($^\dagger$) and significantly higher than ($^*$) the corresponding results using \gls{malpem} are also indicated.}
\begin{small}
\begin{tabular}{@{}p{0.35cm}C{0.38cm}C{0.38cm}C{0.4cm}C{0.38cm}C{0.38cm}C{0.38cm}C{0.38cm}C{0.4cm}C{0.4cm}C{0.42cm}}
\toprule
                  & \multicolumn{5}{c}{Accuracy}                            & \multicolumn{5}{c}{AUC}                                 \\ \midrule
\textit{N}  & 1    & 3    & 6    & 12   & 24                  & 1    & 3    & 6    & 12   & 24                   \\
BL          & 64.3 & 62.9 & 66.0 & 63.0 & 63.3 & 56.4 & 47.9 & 55.0 & 58.7$^\dagger$ & 57.6$^\dagger$ \\
Aug. & \textbf{66.0} & \textbf{64.6} & \textbf{67.8}$^\dagger$ & 62.6 & \textbf{66.6}                       & 57.0 & \textbf{56.7} & \textbf{65.6}$^*$ & \textbf{60.0}$^*$ & \textbf{66.7}$^*$  \\ \midrule
M & & & 67.6 & & &  & & 58.5 &  &\\ \bottomrule
\end{tabular}
\end{small}
\end{figure}

\begin{figure*}[h!]
  \centering
  \includegraphics[trim={7cm 0cm 6.5cm 0cm},clip,width=1\linewidth]{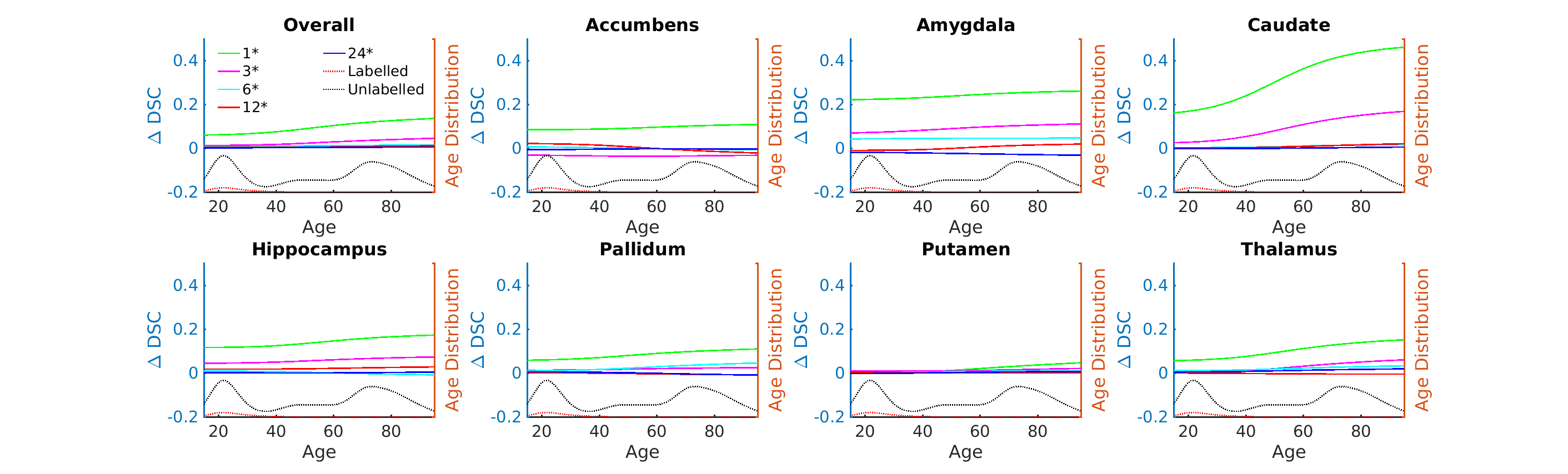}
\caption{Difference in \gls{dsc} using \gls{malpem} segmentations for reference, overall and for each structure, at different ages. Pairwise differences in observed \gls{dsc} between segmentations computed with and without synthetic data augmentation for each level of available labelled images (1,3,6,12,24) are shown. The relative age distributions for the labelled and unlabelled dataset are also shown in each figure. All data is smoothed using kernel regression to highlight the overall trends.}
\label{fig:AgeDelta}
\end{figure*}

\begin{figure*}[h!]
  \centering
  \includegraphics[trim={27cm 12cm 24cm 0cm},clip,width=1\linewidth]{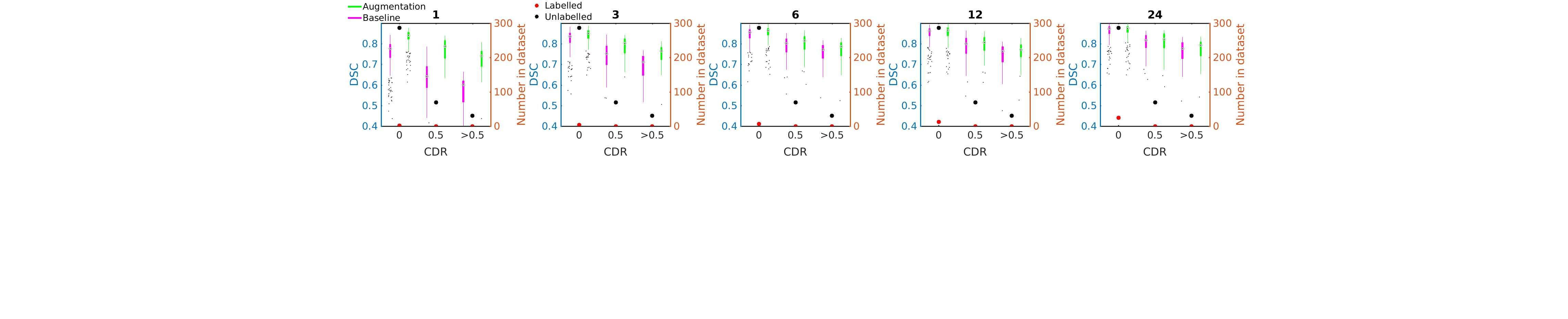}
\caption{Distribution of overall \gls{dsc} using \gls{malpem} segmentations for reference for subjects with different \gls{cdr} levels. The number of each group within the labelled and unlabelled datasets are also indicated.}
\label{fig:CDROverall}
\end{figure*}

Figure~\ref{fig:AgeDelta} shows how the expected improvement in \gls{dsc} varies with age across each structure. In each case there is a clear trend showing that the further away from the ages present in the labelled dataset, the more benefit is given by using the synthetic data. This is particularly noticeable in the caudate. The proximity of the caudate to the lateral ventricles means that its location can vary significantly as the ventricles become enlarged by age, even if its volume is generally preserved. It is therefore a structure which could suffer from a spatial bias being learned when only a few healthy examples are provided, and thus is a particular beneficiary of the additional anatomical variation introduced by the unlabelled data.

\begin{figure*}[h!]
  \centering
  \includegraphics[trim={8cm 0cm 7.5cm 0cm},clip,width=1\linewidth]{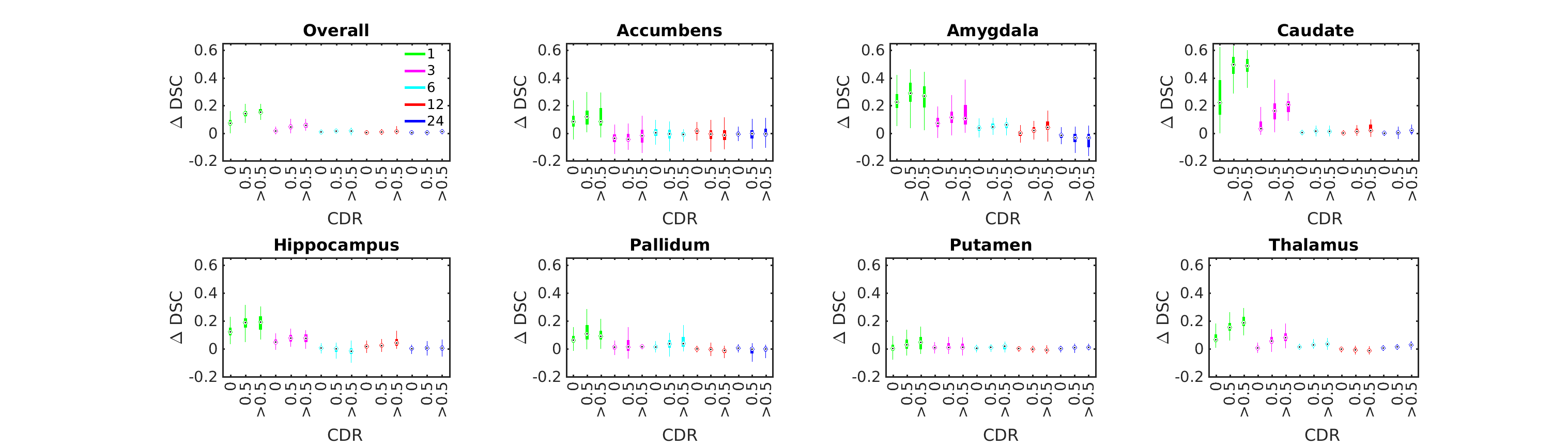}
\caption{Distribution of paired DSC differences between results with and without synthetic data augmentation using \gls{malpem} segmentations for reference. Overall results and results for each structure are shown for subjects with different \gls{cdr} levels. Outliers are omitted for clarity.}
\label{fig:CDRRegions}
\end{figure*}

Figures~\ref{fig:CDROverall} and~\ref{fig:CDRRegions} show a similar analysis, using \gls{cdr} in place of age. Once again, the labelled dataset contains no examples of \gls{ad} pathology, while the full dataset contains significantly more. Again we observe a clear loss of expected segmentation accuracy as \gls{cdr} increases, but this effect is reduced when synthetic data is used. 

The final experiment investigates whether using synthetic data leads to segmentations which were more useful at stratifying cases of \gls{ad}, in particular, distinguishing cases of very mild \gls{ad} (\gls{cdr} 0.5) and mild/moderate \gls{ad} (\gls{cdr} 1/2). Table~\ref{tab:preds} shows a clear benefit to both accuracy and \gls{auc} when synthetic data is used. In fact, using synthetic data increases the accuracy to a level that is close to that achieved by \gls{malpem}, and leads to significantly better results when measured by \gls{auc}.

\subsection{Conclusion}
This paper presented a method for incorporating unlabelled data into a segmentation network. We proposed a novel \gls{gan} training procedure, similar to transfer learning, which allows a \gls{gan} to be trained on a mix of labelled and unlabelled data. The output of this \gls{gan} can then be used to augment real training data. We performed a number of experiments, simulating cases where different levels of labelled data are available, and showed that the proposed method leads to the generation of labelled images with greater anatomical variation than was present in the labelled training data.

Three further sets of experimental results were presented. The first examined the effect of using synthetic data as measured by \gls{dsc} on the labelled dataset. We observed that significant improvements can be made to \gls{dsc} when small amounts of labelled training images are available, particularly in structures which are more strongly affected by \gls{ad}. We also saw that by augmenting with synthetic data, 6 labelled images could be used to achieve the same results as 12 or 24 labelled images. Whilst this provides a useful indicator of performance, it does not evaluate one of the key aims of the proposed method - to extend the domain of training set from young and healthy to old and pathological. To evaluate this, we used a state-of-the-art \gls{3d} multi-atlas segmentation method to generate surrogate ground truth labels for the entire dataset. Using these, we showed how the expected \gls{dsc} changes as age and \gls{ad} diagnosis varies. We observed that segmentation results are substantially reduced once the test images fall outside of the domain of the training images either in terms of age or pathology. This effect was reduced by introducing synthetic data, leading to greater improvements in \gls{dsc} for older subjects and those with \gls{ad}. This confirms our hypothesis that exposing the network to information from unlabelled images of older and more pathological subjects through the use of \gls{gan} derived synthetic data will lead to higher-quality segmentations of such subjects. Finally, the value of these higher-quality segmentations was demonstrated by showing that the stratification of \gls{ad} between very mild and mild or moderate was significantly improved through the use of synthetic data. 

Future work will involve applying the proposed method in different domains and further evaluating its effects. There were some cases where using synthetic data led to worse results, particularly when higher amounts of training data were already available. This suggests the synthetic images are no substitute for real images, implying either a lower image or segmentation quality. Investigating methods to improve image and segmentation quality will also be a subject of future work.
\bibliographystyle{IEEEtran}
\bibliography{local}

\end{document}